\def\year{2020}\relax
\documentclass[letterpaper]{article} 
\usepackage{aaai20}  
\usepackage{times}  
\usepackage{helvet} 
\usepackage{courier}  
\usepackage[hyphens]{url}  
\usepackage{graphicx} 
\usepackage{graphicx}  
\usepackage[separate-uncertainty=true]{siunitx}
\usepackage{amsmath}
\usepackage{amssymb}
\usepackage{physics}
\usepackage[ruled,noline]{algorithm2e}

\usepackage{placeins}

\newcommand{\sub}[1]{_{\mathrm{#1}}}
\newcommand{\etal}{\emph{et al\@.}}
\newcommand{\ie}{i.e.\@}

\newcommand{\relu}{\mathrm{relu}}
\newcommand{\dash}{^{\prime}}
\newcommand{\mathhyph}{{\operatorname{-}}} 
\newcommand{\pmstack}[2]{\substack{+#1 \\ -#2}}
\newcommand{\tblmath}[1]{$#1$}
\newcommand{\noentry}{---}

\usepackage{mathtools,eqparbox}
\newcommand{\indices}[2]{{
  \begin{array}{@{}r@{}}
    \scriptstyle #2~\smash{\eqmakebox[ind]{$\scriptstyle\rightarrow$}} \\[-\jot]  
    \scriptstyle #1~\smash{\eqmakebox[ind]{$\scriptstyle\downarrow$}}
  \end{array}}}
  
\usepackage{color, colortbl}
\definecolor{Gray}{gray}{0.9}

\newcommand{\State}{\mathcal{S}}
\newcommand{\Action}{\mathcal{A}}
\newcommand{\Transition}{\mathcal{T}}
\newcommand{\Reward}{\mathcal{R}}
\newcommand{\Policy}{\pi}
\newcommand{\R}{\mathbb{R}}
\newcommand{\Expected}{\mathbb{E}}

\title{Exploratory Combinatorial Optimization with Reinforcement Learning}
\author{
Thomas D. Barrett,\textsuperscript{\rm 1}
William R. Clements,\textsuperscript{\rm 2}
Jakob N. Foerster,\textsuperscript{\rm 3}
Alex I. Lvovsky\textsuperscript{\rm 1,4}\\
\textsuperscript{\rm 1}University of Oxford, Oxford, UK\\
\textsuperscript{\rm 2}indust.ai, Paris, France\\
\textsuperscript{\rm 3}Facebook AI Research\\
\textsuperscript{\rm 4}Russian Quantum Center, Moscow, Russia\\
thomas.barrett@physics.ox.ac.uk,
william.clements@indust.ai,
jnf@fb.com,
alex.lvovsky@physics.ox.ac.uk
}

\begin{document}

\maketitle

\begin{abstract}

Many real-world problems can be reduced to combinatorial optimization on a graph, where the subset or ordering of vertices that maximize some objective function must be found.  With such tasks often NP-hard and analytically intractable, reinforcement learning (RL) has shown promise as a framework with which efficient heuristic methods to tackle these problems can be learned.  Previous works construct the solution subset incrementally, adding one element at a time, however, the irreversible nature of this approach prevents the agent from revising its earlier decisions, which may be necessary given the complexity of the optimization task. We instead propose that the agent should seek to continuously improve the solution by learning to \emph{explore at test time}.  Our approach of exploratory combinatorial optimization (ECO-DQN) is, in principle, applicable to any combinatorial problem that can be defined on a graph. Experimentally, we show our method to produce state-of-the-art RL performance on the Maximum Cut problem.  Moreover, because ECO-DQN can start from any arbitrary configuration, it can be combined with other search methods to further improve performance, which we demonstrate using a simple random search.

\end{abstract}

\section{Introduction}

NP-hard combinatorial problems -- such as Travelling Salesman~\cite{papadimitriou77}, Minimum Vertex Cover~\cite{dinur05} and Maximum Cut~\cite{goemans95} -- are canonical challenges in computer science.  With practical applications ranging from fundamental science to industry, efficient methods for approaching combinatorial optimization are of great interest.  However, as no known algorithms are able to solve NP-hard problems in polynomial time, exact methods rapidly become intractable.  Approximation algorithms guarantee a worst-case solution quality, but sufficiently strong bounds may not exist and, even if they do, these algorithms can have limited scalability~\cite{williamson11}.  Instead, heuristics are often deployed that, despite offering no theoretical guarantees, are chosen for high performance.

There are numerous heuristic methods, ranging from search-based~\cite{benlic13,banks08} to physical systems that utilise both quantum and classical effects~\cite{johnson11,yamamoto17} and their simulated counterparts~\cite{kirkpatrick83,clements17,tiunov19}.  However, the effectiveness of general algorithms is dependent on the problem being considered, and high levels of performance often require extensive tailoring and domain-specific knowledge.  Machine learning offers a route to addressing these challenges, which led to the demonstration of a meta-algorithm, S2V-DQN~\cite{khalil17}, that utilises reinforcement learning (RL) and a deep graph network to automatically learn good heuristics for various combinatorial problems.

A solution to a combinatorial problem defined on a graph consists of a subset of vertices that satisfies the desired optimality criteria.  Approaches following S2V-DQN's framework incrementally construct solutions one element at a time -- reducing the problem to predicting the value of adding any vertex not currently in the solution to this subset.  However, due to the inherent complexity of many combinatorial problems, learning a policy that directly produces a single, optimal solution is often impractical, as evidenced by the sub-optimal performance of such approaches.  Instead, we propose that a natural reformulation is for the agent to explore the solution space at test time, rather than producing only a single ``best-guess''.  Concretely, this means the agent can add or remove vertices from the solution subset and is tasked with searching for ever-improving solutions at test time.  In this work we present ECO-DQN (Exploratory Combinatorial Optimization DQN), a framework combining RL and deep graph networks to realise this approach.

Our experimental work considers the Maximum Cut (Max-Cut) problem as it is a fundamental combinatorial challenge -- in fact over half of the 21 NP-complete problems enumerated in Karp's seminal work~\cite{karp72} can be reduced to Max-Cut -- with numerous real-world applications.  The framework we propose can, however, be readily applied to any graph-based combinatorial problem where solutions correspond to a subset of vertices and the goal is to optimize some objective function.

By comparing ECO-DQN to S2V-DQN as a baseline, we demonstrate that our approach improves on the state-of-the-art for applying RL to the Max-Cut problem.  Suitable ablations show that this performance gap is dependent on both allowing the agent to reverse its earlier decisions and providing suitable information and rewards to exploit this freedom.  Moreover, ECO-DQN can be initialised in any state (\ie{} will look to improve on any proposed solution) and, as a consequence, has the flexibility to be either deployed independently or combined with other search heuristics.  For example, we achieve significant performance improvements by simply taking the best solution found across multiple randomly initialised episodes.  ECO-DQN also generalises well to graphs from unseen distributions.  We obtain very strong performance on known benchmarks of up to 2000 vertices, even when the agent is trained on graphs an order of magnitude smaller and with a different structure. 

\subsection{Related Work}

A formative demonstration of neural networks for combinatorial optimization (CO) was the application of Hopfield networks to the Travelling Salesman Problem (TSP) by Hopfield and Tank~\shortcite{hopfield85}.  They mapped $N$-city problems to $N{\times}N$ graphs (networks) with each vertex, $v_{ij}$,  a binary variable denoting whether city $i$ is the $j$-th to be visited, and the edges connecting vertices proportional to the distance between cities.  Although the results of Hopfield and Tank were contested by Wilson and Pawley~\shortcite{wilson88}, there followed a period active research into neural networks for CO that lasted for over a decade \cite{smith99}.  During this time, RL techniques were first by applied to CO by Zhang and Diettench \shortcite{zhang95}, who considerer the NP-hard job-shop problem (of which the TSP is a specific case).

More recently, Bello~\etal{}~\shortcite{bello16} used policy gradients to train pointer networks \cite{vinyals15}, a recurrent architecture that produces a softmax attention mechanism (a ``pointer'') to select a member of the input sequence as an output. However, this architecture did not reflect the structure of problems defined over a graph, which Khalil \etal{}~\shortcite{khalil17} addressed with S2V-DQN, a general RL-based framework for CO that uses a combined graph embedding network and deep Q-network. Mittal \etal{} \shortcite{mittal19} developed these ideas further by modifying the  training process: first training an embedding graph convolution network (GCN), and then training a Q-network to predict the vertex (action) values.  This is orthogonal to our proposal which considers the framework itself, rather than the training procedure, and, in principle, appears to be compatible with ECO-DQN.  

Another current direction is applying graph networks for CO in combination with a tree search.  Li \etal{}~\shortcite{li18} combined a GCN with a guided tree-search in a supervised setting, \ie{} requiring large numbers of pre-solved instances for training.  Very recently, Abe \etal{}~\shortcite{abe19} trained a GCN using Monte-Carlo tree search as a policy improvement operator, in a similar manner to AlphaGo Zero~\cite{silver17}, however, this work does not consider the Max-Cut problem.

\section{Background}

\subsection{Max-Cut Problem}

The Max-Cut problem is to find a subset of vertices on a graph that maximises the total number of edges connecting vertices within this subset to vertices not in this subset (the cut value).  In this work we consider the more general weighted version of this problem, where each edge in the graph is assigned a weight and the objective is to maximise the total value of cut edges.  Formally, for a graph, $G(V,W)$, with vertices $V$ connected by edges $W$, the Max-Cut problem is to find the subset of vertices $S \subset V$ that maximises $C(S,G)=\sum_{i{\subset}S,j{\subset}V{\setminus}S} w_{ij}$ where $w_{ij}\in W$ is the weight of the edge connecting vertices $i$ and $j$.

This is not simply a mathematical challenge as many real world applications can be reduced to the Max-Cut problem, including protein folding~\cite{perdomo12}, investment portfolio optimization~\cite{elsokkary17,venturelli18} (specifically using the Markowitz~\shortcite{markowitz52} formulation), and finding the ground state of the Ising Hamiltonian in physics~\cite{barahona82}.

\subsection{Q-learning}

As is standard for RL, we consider the optimization task as a Markov decision process (MDP) defined by the 5-tuple $(\State,\Action,\Transition,\Reward,\gamma)$.  Here, $\State$ denotes the set of states, $\Action$ is the set of actions, $\Transition : \State \times \Action \times \State \rightarrow [0,1]$ is the transition function, $\Reward : \State \rightarrow \R$ the reward function and $\gamma \in [0,1]$ is the discount factor.  A policy, $\pi :  \State \rightarrow [0,1]$, maps a state to a probability distribution over actions.  The Q-value of a given state-action pair, $(s \in \State, a \in \Action)$, is then given by the discounted sum of immediate and future rewards
\begin{equation}
	Q^{\Policy}(s,a) = \Expected\biggl[ \sum_{t=0}^{\infty} \gamma^{t} \Reward(s_t) \Bigm| s_0 = s, a_0=a, \Policy \biggr],
\end{equation}
where $s_0$ and $a_0$ correspond to the initial state and action taken, with future actions chosen according to the policy, $\Policy$.  

A deep Q-network~\cite{mnih15} (DQN) provides a function $Q(s,a;\theta)$, where $\theta$ parameterises the network, which is trained to approximate ${Q^{*}(s,a) \equiv \mathrm{max}_{\Policy}Q^{\pi}(s,a)}$, the Q-values of each state-action pair when following the optimal policy.  Once trained, an approximation of the optimal policy can be obtained simply by acting greedily with respect to the predicted Q-values, ${ \pi(s;\theta) = \mathrm{argmax}_{a\dash} Q(s,a\dash;\theta)}$.

\subsection{Message Passing Neural Networks}

Our choice of deep Q-network is a message passing neural network (MPNN)~\cite{gilmer17}.  This is a general framework of which many common graph networks are specific implementations.  Each vertex in the graph, $v \in V$, is represented with an $n$-dimensional embedding, $\mu_{v}^{k}$, where $k$ labels the current iteration (network layer).  These are initialised, by some function $I$, from an input vector of observations, $x_{v}$, as $\mu_{v}^{0} = I(x_{v})$.  During the message-passing phase, the embeddings are repeatedly updated with information from neighbouring vertices, $N(v)$, according to
\begin{align}
	m_{v}^{k+1} &= M_{k}\big(\mu_{v}^k, \{\mu_{u}^k\}_{u {\in} N(v)}, \{w_{uv}\}_{u {\in} N(v)}\big), \\
	\mu_{v}^{k+1} &= U_{k}\big(\mu_{v}^k,m_{v}^{k+1} \big),
\end{align}
where $M_{k}$ and $U_{k}$ are message and update functions, respectively.  After $K$ rounds of message passing, a prediction is produced by some readout function, $R$. In our case this prediction is the set of Q-values of the actions corresponding to ``flipping" each vertex, i.e.~adding or removing it from the solution subset $S$, 	$\{Q_{v}\}_{v\in V} = R ( \{\mu_{u}^{K}\}_{u {\in} V} )$.

\section{Exploiting Exploration}
\label{sec:exploitingExploration}

One straightforward application of Q-learning to CO over a graph is to attempt to directly learn the utility of adding any given vertex to the solution subset.  This formalism, which is followed by S2V-DQN and related works, incrementally constructs a solution by adding one vertex at a time to the subset, until no further improvement can be made.  However, the complexity of NP-hard combinatorial problems means it is challenging to learn a single function approximation of $Q^{*}(s,a)$ that generalises across the vast number of possible graphs.  Therefore, as vertices can only be added to the solution set, policies derived from the learnt Q-values, such as a typical greedy policy, will likely be sub-optimal.

In this work we present an alternative approach where the agent is trained to explore the solution space at test time, seeking ever-improving states.  As such, the Q-value of either adding or removing a vertex from the solution is continually re-evaluated in the context of the episode's history.  Additionally, as all actions can be reversed, the challenge of predicting the true value of a vertex ``flip" does not necessarily result in sub-optimal performance.  The fundamental change distinguishing our approach, ECO-DQN, from previous works can then be summarised as follows: \emph{instead of learning to construct a single good solution, learn to explore for improving solutions}.

However, simply allowing for revisiting the previously flipped vertices does not automatically improve performance.  The agent is not immediately able to make more informed decisions, nor can it reach previously unobtainable solutions.  Instead, further modifications are required to leverage this freedom for improved performance, which we now discuss.

\subsubsection{Reward Shaping.}

The objective of our exploring agent is to find the best solution (highest cut-value) at any point within an episode.  Formally, the reward at state $s_{t} \in \State$ is given by ${\Reward(s_{t}) = \mathrm{max}(C(s_{t}) - C(s^{*}),\  0) / \abs{V}}$, where $s^{*} \in \State$ is the state corresponding to the highest cut value previously seen within the episode, $C(s^{*})$ (note that we implicitly assume the graph, $G$, and solution subset, $S$, to be included in the state).  As continued exploration is desired, even after a good solution is found, there is no punishment if a chosen action reduces the cut-value.  The reward is normalised by the total number of vertices, $\abs{V}$, to mitigate the impact of different reward scales across different graph sizes.  We use a discount factor of $\gamma=0.95$ to ensure the agent actively pursues rewards within a finite time horizon.

As our environment only provides a reward when a new best solution is found, after an initial period of exploration, these extrinsic rewards can be sparse, or absent, for the remainder of the episode.  We therefore also provide a small intermediate reward of $1/\abs{V}$ whenever the agent reaches a locally optimal state (one where no action will immediately increase the cut value) previously unseen within the episode.  In addition to mitigating the effect of sparse extrinsic rewards, these intrinsic rewards also shape the exploratory behaviour at test time. There are far more states than could be visited within our finite episodes, the vast majority of which are significantly sub-optimal, and so it is useful to focus on a subset of states known to include the global optimum.  As local optima in combinatorial problems are typically close to each other, the agent learns to ``hop'' between nearby local optima, thereby performing a in-depth  local search of the most promising subspace of the state space (see figure \ref{fig:intraepisode_behaviour}c).

\subsubsection{Observations}

A Q-value for flipping each vertex is calculated using seven observations, ($x_{v} {\in} \R^{7}$), derived from the current state, with a state corresponding to both the target graph, $G(V,W)$, and the current subset of vertices assigned to the solution set, $S \subset V$.  These observations are:
	\begin{enumerate}
		\item Vertex state, \ie{} if $v$ is currently in the solution set, $S$.
		\item Immediate cut change if vertex state is changed.
		\item Steps since the vertex state was last changed.
		\item Difference of current cut-value from the best observed.
		\item Distance of current solution set from the best observed.
		\item Number of available actions that immediately increase the cut-value.
		\item Steps remaining in the episode.
	\end{enumerate}
Observations (1-3) are local, which is to say they can be different for each vertex considered, whereas (4-7) are global, describing the overall state of the graph and the context of the episode.  The general purposes of each of the observations are: (1-2) provide useful information for determining the value of selecting an action; (3) provides a simple history to prevent short looping trajectories; (4-6) ensure the extrinsic and intrinsic rewards are Markovian; and (7) accounts for the finite episode duration.

\subsection{Experiments}

\begin{figure*}[!t]
\centering
\includegraphics{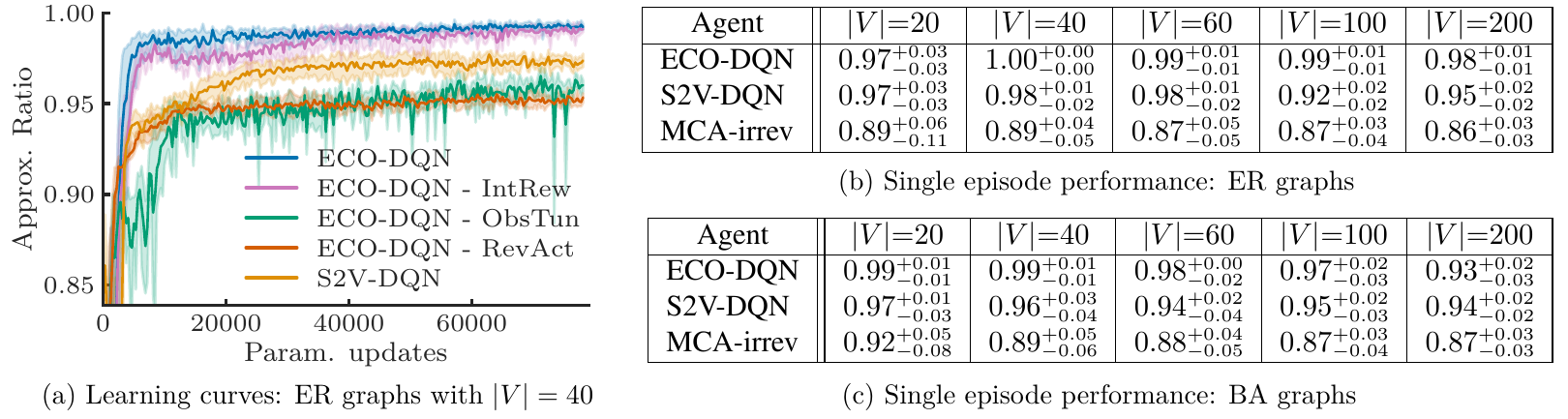}
\caption{
Performance comparison of ECO-DQN and baselines.
(a) Learning curves, averaged over 5 seeds, when training on 40-vertex ER graphs.
(b-c) Approximation ratios for ER and BA graphs with different numbers of vertices, $\abs{V}$.  We report the mean approximation ratios over the 100 validation graphs, along with the distance to the upper and lower quartiles.
\label{fig:performance_benchmarking}
}
\end{figure*}

\subsubsection{Experimental details.}

In this work we train and test the agent on both Erd\H{o}s-R\'{e}nyi ~\cite{erdos60} and Barabasi-Albert ~\cite{albert02} graphs with edges $w_{ij}\in\{0,{\pm}1\}$, which we refer to as ER and BA graphs, respectively.  Training is performed on randomly generated graphs from either distribution, with each episode considering a freshly generated instance.  The performance over training (\ie{} all learning curves) is evaluated as the mean of a fixed set of 50 held-out graphs from the same distribution.  Once trained, the agents are tested on a separate set of 100 held-out validation graphs from a given distribution.

During training and testing, every action taken demarks a timestep, $t$.  For agents that are allowed to take the same action multiple times (\ie{} ECO-DQN and selected ablations), which for convenience we will refer to as \emph{reversible} agents, the episode lengths are set to twice the number of vertices in the graph, $t = 1, 2, \dots, 2\abs{V}$. Each episode for such agents is initialised with a random subset of vertices in the solution set.  By contrast, agents that can only add vertices to the solution set (\emph{irreversible} agents, \ie{} S2V-DQN and selected ablations) are initialised with an empty solution subset. These agents keep selecting actions greedily even if no positive Q-values are available until $t=|V|$, to account for possible incorrect predictions of the Q-values.  In both cases the best solution obtained at any timestep within the episode is taken as the final result.  To facilitate direct comparison, ECO-DQN and S2V-DQN are implemented with the same MPNN architecture, with details provided in the Appendix.

\subsubsection{Benchmarking details}

We compare the performance of ECO-DQN to a leading RL-based heuristic, S2V-DQN.  To interpret the performance gap, we also consider the following ablations, which together fully account for the differences between our approach and the baseline ($\mathrm{ECO\mathhyph{}DQN} \equiv \mathrm{S2V\mathhyph{}DQN} + \mathrm{RevAct} + \mathrm{ObsTun} + \mathrm{IntRew}$).

\begin{itemize}
	\item \emph{Reversible Actions} (RevAct): Whether the agent is allowed to flip a vertex more than once.  For irreversible agents we follow S2V-DQN and use $\gamma{=}1$.
	\item \emph{Observation Tuning} (ObsTun): Observations (2-7) from the list above that allow the agent to exploit having reversible actions.  Also, in the absence of ObsTun, the rewards used are simply the (normalised) immediate change in cut value, $\Reward(s_t) = (C(s_t) - C(s_{t-1}))  / \abs{V}$, which is necessary as without observations (4-5) the ECO-DQN reward structure is non-Markovian.
	\item \emph{Intermediate Rewards} (IntRew): Whether the agent is provided with the small intermediate rewards for reaching new locally optimal solutions.
\end{itemize}

As an additional benchmark we also implement the MaxCutApprox (MCA) algorithm.  This is a greedy algorithm, choosing the action (vertex) that provides the greatest immediate increase in cut value until no further improvements can be made. We consider two modifications of MCA. The standard application, which we denote MCA-irrev, is irreversible and begins with an empty solution set. The alternative algorithm, MCA-rev, starts with a random solution set and allows reversible actions.

We use the approximation ratio -- ${C(s^{*})/C(s\sub{opt})}$, where $C(s\sub{opt})$ is the cut-value of the true optimum solution -- of each approach as a metric of solution quality.  Exact methods are intractable for many of the graph sizes we use, therefore we apply a battery of optimization approaches to each graph and take the best solution found by any of them as the ``optimum'' solution.  Specifically, in addition to ECO-DQN, S2V-DQN and the MCA algorithms, we use CPLEX, an industry standard integer programming solver, and a pair of recently developed simulated annealing heuristics by Tiunov \etal~\shortcite{tiunov19} and Leleu \etal~\shortcite{leleu19}.  Details of these implementations and a comparison of their efficacy can be found in the Supplemental Material.

\subsubsection{Performance benchmarking.} 

\begin{figure*}[!t]
\centering
\includegraphics{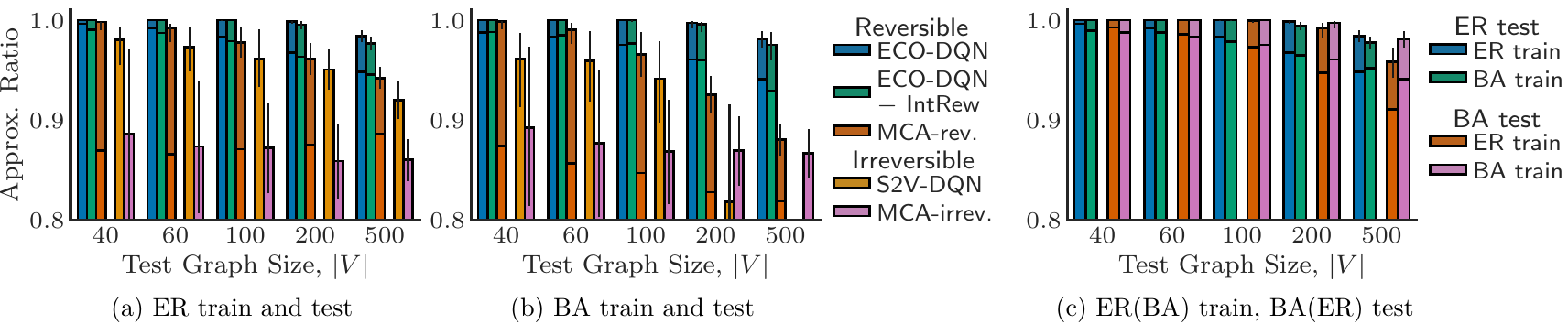}
\caption{Generalisation of agents to unseen graph sizes and structures.  (a-b)~The performance of agents trained on ER and BA graphs of size $\abs{V}{=}40$ tested on graphs of up to $\abs{V}{=}500$ of the same type.  ECO-DQN is shown with and without providing the intermediate rewards (IntRew) for finding locally optimal solutions during training, and is compared to the MCA-rev algorithm.  Reversible agents are applied with 50 randomly initialised episodes to each of the 100 validation graphs for each type and size.  The first marking on each bar is the average across every episode (the expected `single-try' performance), with the upper limit extending to the average performance across different graphs.  The vertical bars denote the \SI{68}{\percent} confidence interval of this upper limit.  Irreversible approaches are initialised with empty solution sets, and so only use 1 episode per graph.  Note that S2V-DQN applied to BA graphs with $\abs{V}{=}500$ is not visible on these axes but has a value of $0.49\pmstack{0.12}{0.11}$.  (c)~A comparison of how agents trained on only one of either ER or BA graphs with $\abs{V}{=}40$, perform on larger graphs from both distributions.}
\label{fig:agent_generalisation}
\end{figure*}

Figure \ref{fig:performance_benchmarking}a shows learning curves of agents trained on ER graphs of size $\abs{V}=40$, where it can be seen that ECO-DQN reaches a significantly higher average cut than S2V-DQN.  Removing either reversible actions (RevAct) or the additional observations (ObsTun) reduces the performance below that of S2V-DQN, underlining our previous assertion that obtaining state-of-the-art performance requires not only that the agent be allowed to reverse its previous actions, but also that it be suitably informed and rewarded to do so effectively.  Intermediate rewards (IntRew) are seen to speed up and stabilise training.  They also result in a small performance improvement, however this effect becomes clearer when considering how the agents generalise to larger graphs (see figures \ref{fig:agent_generalisation}a and \ref{fig:agent_generalisation}b).

Figures \ref{fig:performance_benchmarking}b and \ref{fig:performance_benchmarking}c show the performance of agents trained and tested on graphs with up to 200 vertices.  We see that ECO-DQN has superior performance across most considered graph sizes and structures.  Both ECO-DQN and S2V-DQN have similar computational costs per action, with extended performance comparisons provided in the appendix.

\subsubsection{Intra-episode behaviour.}

\begin{figure}[t!]
\centering
\includegraphics{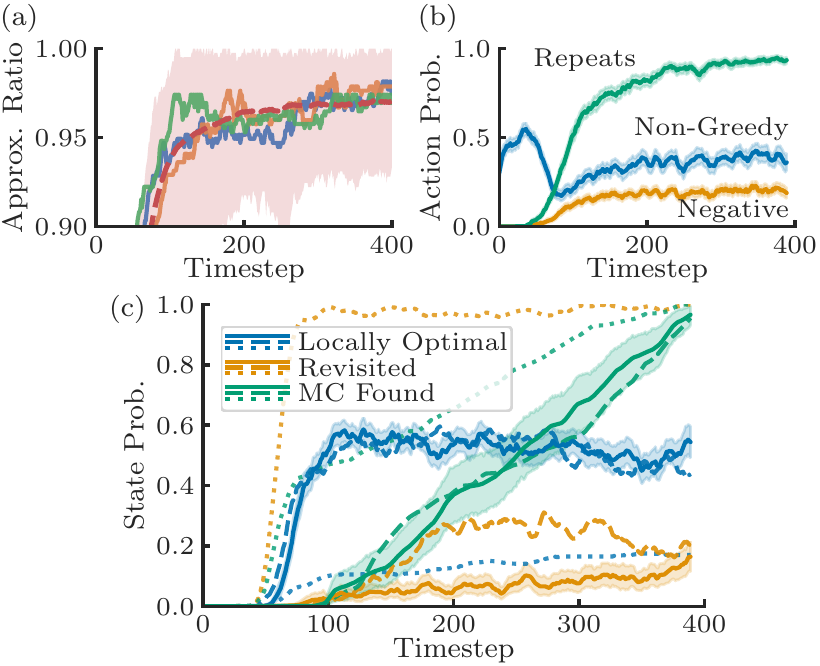}
\caption{
Intra-episode behaviour averaged across all 100 instances from the validation set for ER graphs with $\abs{V}=200$.
(a)~Mean (dashed) and range (shaded) of all trajectories, with three examples (solid) shown for reference.
(b)~The probability that the chosen action has already been taken within the episode (Repeats), does not provide the greatest immediate reward (Non-Greedy) or reduces the cut-value (Negative).
(c)~The probability that the current state is locally optimal (Locally Optimal), has already been visited within the episode (Revisited), and that the best solution that will be found within the episode has already been seen (MC found).
The behaviour is shown at three points during training: when performance is equivalent to that of MCA-irrev (dotted) or S2V-DQN (dashed), and when fully trained (solid). 
(b-c) use a 10-step moving average over all graphs (trajectories) in the validation set, however, the shaded errors are only shown for the fully trained agent in (c).
}
\label{fig:intraepisode_behaviour}
\end{figure}

We now consider how this strong performance is achieved by examining the intra-episode behaviour of an agent trained and tested on 200-vertex ER graphs.  The larger graph size is chosen as it provides greater scope for the agent to exhibit non-trivial behaviour.  Figure \ref{fig:intraepisode_behaviour}a highlights the trajectories taken by the trained agent on graphs from the validation set. Whilst the overall trend is towards higher cut-values, the fluctuations show that the agent has learnt to search for improving solutions even when this requires sacrificing cut-value in the short-term.  Further analysis of the agent's behaviour is presented in figures \ref{fig:intraepisode_behaviour}b and \ref{fig:intraepisode_behaviour}c which show the action preferences and the types of states visited, respectively, over the course of an optimization episode.

From figure \ref{fig:intraepisode_behaviour}b, we see that the fully trained agent regularly chooses actions that do not correspond to the greatest immediate increase in the cut-value (Non-Greedy), or even that decrease the cut value (Negative).  Moreover, the agent also moves the same vertex in or out of the solution set multiple times within an episode (Repeats), which suggests the agent has learnt to explore multiple possible solutions that may be different from those obtained initially.

This is further emphasised in figure \ref{fig:intraepisode_behaviour}c where we see that, after an initial period of exploration, the agent searches through the solution space, repeatedly moving in and out of locally optimal (Locally Optimal) solutions whilst minimising the probability that it revisits states (Revisited).  By comparing the agent at three points during training (fully trained and when performance level is equivalent to either MCA-irrev or S2V-DQN), we see that this behaviour is learnt.  Weaker agents from earlier in training revisit the same states far more often, yet find fewer locally optimal states.  The probability that the fully-trained agent has already found the best solution it will see in the episode (MC found) grows monotonically, implying that the agent finds ever better solutions while exploring.  Indeed, for this agent, simply increasing the number of timesteps in an episode from $2\abs{V}{=}400$ to $4\abs{V}$ is seen to increase the average approximation ratio from $0.98\pmstack{0.01}{0.01}$ to $0.99\pmstack{0.01}{0.01}$.  The range quoted for these approximation ratios corresponds to the upper and lower quartiles of the performance across all 100 validation graphs.

\section{Leveraging Variance}
\label{sec:leveraging_varience}

Changing the initial subset of vertices selected to be in the solution set can result in very different trajectories over the course of an episode.  An immediate result of this stochasticity is that performance can be further improved by running multiple episodes with a distribution of initialisations, and selecting the best result from across this set.

\subsection{Experiments}

We optimize every graph using 50 randomly initialised episodes.  At the same time, we make the task more challenging by testing on graphs that are larger, or that have a different structure, from those on which the agent was trained.  This ability to generalise to unseen challenges is important for the real-world applicability of RL agents to combinatorial problems where the distribution of optimization tasks may be unknown or even change over time.

\subsubsection{Generalisation to unseen graph types.}

Figures \ref{fig:agent_generalisation}a and \ref{fig:agent_generalisation}b show the generalisation of agents trained on 40 vertices to systems with up to 500 vertices for ER and BA graphs, respectively.  (Generalisation data for agents trained on graphs of sizes ranging from $\abs{V}=20$ to $\abs{V}=200$ can be found in the Appendix.)  ECO-DQN is compared to multiple benchmarks, with details provided in the caption, however there are three important observations to emphasise.  Firstly, reversible agents outperform the irreversible benchmarks on all tests, with the performance gap widening with increasing graph size.  This is particularly noticeable for BA graphs, for which the degrees of each vertex in the graph tend to be distributed over a greater range than for ER graphs, where S2V-DQN fails to generalise in any meaningful way to $\abs{V}{\geq}200$.

Secondly, for the reversible agents it is clear that using multiple randomly initialised episodes provides a significant advantage.  As ECO-DQN provides near-optimal solutions on small graphs within a single episode, it is only on larger graphs that this becomes relevant.  However, it is noteworthy that even the simple MCA-rev algorithm, with only a relatively modest budget of 50 random initialisations, outperforms a highly trained irreversible heuristic (S2V-DQN).  This further emphasises how stochasticity -- which here is provided by the random episode initialisations and ensures many regions of the solution space are considered -- is a powerful attribute when combined with local optimization.

Finally, we again observe the effect of small intermediate rewards (IntRew) for finding locally optimal solutions during training upon the final performance.  For small graphs the agent performs near-optimally with or without this intrinsic motivation, however the difference becomes noticeable when generalising to larger graphs at test time.

In figure \ref{fig:agent_generalisation}c we observe that ECO-DQN performs well across a range of graph structures, even if they were not represented during training, which is a highly desirable characteristic for practical CO. We train the agent on ER graphs with $\abs{V}{=}40$ and then test it on BA graphs of up to $\abs{V}{=}500$, and vice versa. The performance is marginally better when testing on graphs from the same distribution as the training data, however this difference is negligible for $\abs{V} \leq 100$.  Furthermore, in every case deploying ECO-DQN with 50 randomly initialised episodes outperforms all other benchmarks (S2V-DQN and MCA), even when only ECO-DQN is trained on different graph types to the test data.

\subsubsection{Generalization to real-world datasets.}

Finally, we test ECO-DQN on publicly available datasets.  The ``Physics'' dataset consists of ten graphs -- with $\abs{V}{=}125$, exactly 6 connections per vertex and $w_{ij}{\in}\{0,{\pm}1\}$ -- corresponding to Ising models of physical systems.  The GSet is a well-investigated benchmark collection of graphs~\cite{benlic13}.  We separately consider ten graphs, G1-G10, with $\abs{V}{=}800$, and ten larger graphs,  G22-G32, with $\abs{V}{=}2000$.  For G1-G10 we utilise 50 randomly initialised episodes per graph, however for G22-G32 we use only a single episode per graph, due to the increased computational cost.  We apply agents trained on ER graphs with $\abs{V}{=}200$. The results are summarised in table \ref{tab:real_world_benchmarks}, where ECO-DQN is seen to significantly outperform other approaches, even when restricted to use only a single episode per graph.

\begin{table}[!ht]
\centering
\begin{tabular}{|l||c|c|c|}
\hline
Dataset & ECO-DQN & S2V-DQN & MCA-(rev,\,irrev) \\
\hline
Physics & 1.000 & 0.928 & 0.879, 0.855 \\
G1-10 & 0.996 & 0.950 & 0.947, 0.913 \\
G22-32 & 0.971 & 0.919 & 0.883, 0.893 \\
\hline
\end{tabular}
\caption{Average performance on known benchmarks.}
\label{tab:real_world_benchmarks}
\end{table}


Despite the structure of graphs in the ``Physics'' dataset being distinct from the ER graphs on which the agent is trained, every instance in optimally solved.  Averaged across all graphs, \SI{37.6}{\percent} of episodes find an optimal solution and \SI{90.4}{\percent} of these solutions are unique, demonstrating that, in conjunction with random initialisations, the agent is capable of finding many different optimal trajectories.  Importantly, the structure of the GSet is distinct from that of the training data, with the first five instances in each tested set have only positive edges, ${w_{ij} {\in} \{0,1\}}$.

\section{Summary and Outlook}

This work introduces ECO-DQN, a new state-of-the-art RL-based algorithm for the Max-Cut problem that generalises well to unseen graph sizes and structures.  We show that treating CO as an ongoing exploratory exercise in surpassing the best observed solution is a powerful approach to this NP-hard problem.  In principle, our approach is applicable to any combinatorial problem defined on a graph.

ECO-DQN can initialise a search from any valid state, opening the door to combining it with other search heuristics. We obtained further improved performance with a simple ``heuristic" of randomly initialised episodes, however, one could consider combining ECO-DQN with more sophisticated episode initialisation policies. Alternatively, ECO-DQN could also be initialised with solutions found by other optimization methods to further strengthen them.  Also, we train our agents with highly discounted future rewards ($\gamma{=0.95}$), and although this is found to provide strong performance, the relatively short-term reward horizon likely limits exploration to only local regions of the solution space.  As such, it would be interesting to investigate longer reward-horizons, particularly when training on larger graphs.  A more substantial avenue to explore would be to use a recurrent architecture to learn a useful representation of the episode history, as opposed to the hand-crafted representation that we describe in section \ref{sec:exploitingExploration}.

Whilst these are paths towards further developing exploration-based CO, we believe that the strong performance already demonstrated would allow our approach to be applied in numerous practical settings.  This is especially true for settings where many graphs of similar structure need to be optimized, such as protein folding~\cite{perdomo12} and portfolio optimization~\cite{elsokkary17,venturelli18}.

\begin{table*}[t!]
\centering
 \resizebox{\textwidth}{!}{
\begin{tabular}{|r|c|c|c|c|c|c||c|c|c|c|c|c|}
\hline
$\indices{\mathrm{Train}}{\mathrm{Test~\,}}$ & \tblmath{\abs{V}{=}20} & \tblmath{\abs{V}{=}40} & \tblmath{\abs{V}{=}60} & \tblmath{\abs{V}{=}100} & \tblmath{\abs{V}{=}200} & \tblmath{\abs{V}{=}500} & \tblmath{\abs{V}{=}20} & \tblmath{\abs{V}{=}40} & \tblmath{\abs{V}{=}60} & \tblmath{\abs{V}{=}100} & \tblmath{\abs{V}{=}200} & \tblmath{\abs{V}{=}500} \\
\hline
& \multicolumn{6}{c||}{ ER graphs} & \multicolumn{6}{c|}{BA graphs} \\
\hline

\tblmath{\abs{V}{=}20} & 
\tblmath{0.99\pmstack{0.01}{0.01}} & \tblmath{1.00\pmstack{0.00}{0.00}} & \tblmath{1.00\pmstack{0.00}{0.00}} & \tblmath{1.00\pmstack{0.00}{0.00}} & \tblmath{0.98\pmstack{0.01}{0.01}} & \tblmath{0.95\pmstack{0.01}{0.01}} & 
\tblmath{1.00\pmstack{0.00}{0.00}} & \tblmath{1.00\pmstack{0.00}{0.00}} & \tblmath{1.00\pmstack{0.00}{0.00}} & \tblmath{1.00\pmstack{0.00}{0.00}} & \tblmath{0.99\pmstack{0.01}{0.01}} & \tblmath{0.98\pmstack{0.01}{0.01}} \\

\tblmath{\abs{V}{=}40} & 
\noentry & \tblmath{1.00\pmstack{0.00}{0.00}} & \tblmath{1.00\pmstack{0.00}{0.00}} & \tblmath{1.00\pmstack{0.00}{0.00}} & \tblmath{1.00\pmstack{0.00}{0.00}} & \tblmath{0.98\pmstack{0.01}{0.01}} & 
\noentry & \tblmath{1.00\pmstack{0.00}{0.00}} & \tblmath{1.00\pmstack{0.00}{0.00}} & \tblmath{1.00\pmstack{0.00}{0.00}} & \tblmath{1.00\pmstack{0.00}{0.00}} & \tblmath{0.98\pmstack{0.01}{0.01}} \\

\tblmath{\abs{V}{=}60} & 
\noentry & \noentry & \tblmath{1.00\pmstack{0.00}{0.00}} & \tblmath{1.00\pmstack{0.00}{0.00}} & \tblmath{1.00\pmstack{0.00}{0.00}} & \tblmath{0.99\pmstack{0.01}{0.01}} &
 \noentry & \noentry & \tblmath{1.00\pmstack{0.00}{0.00}} & \tblmath{1.00\pmstack{0.00}{0.00}} & \tblmath{1.00\pmstack{0.00}{0.00}} & \tblmath{0.99\pmstack{0.01}{0.01}} \\
 
\tblmath{\abs{V}{=}100} & \noentry & \noentry & \noentry & \tblmath{1.00\pmstack{0.00}{0.00}} & \tblmath{1.00\pmstack{0.00}{0.00}} &
\tblmath{1.00\pmstack{0.00}{0.00}} & \noentry & \noentry & \noentry & \tblmath{1.00\pmstack{0.00}{0.00}} & \tblmath{1.00\pmstack{0.00}{0.00}} & \tblmath{0.98\pmstack{0.01}{0.01}} \\

\tblmath{\abs{V}{=}200} & \noentry & \noentry & \noentry & \noentry & \tblmath{1.00\pmstack{0.00}{0.00}} & \tblmath{1.00\pmstack{0.00}{0.00}} & 
\noentry & \noentry & \noentry & \noentry & \tblmath{0.99\pmstack{0.01}{0.01}} & \tblmath{0.98\pmstack{0.01}{0.01}} \\
\hline
\end{tabular}
} 
\caption{Generalisation performance of ECO-DQN, using 50 randomly initialised episodes per graph.}
\label{tab:performance_comps_app}
\end{table*}

\section*{Acknowledgements}

The authors would like to thank D.~Chermoshentsev and A.~Boev.  A.L.'s research is partially supported by Russian Science Foundation (19-71-10092).

\section*{Appendix}

\subsubsection{Code and graph availability.}

Source code, including experimental scripts and all testing and validation graphs, can be found at \url{https://github.com/tomdbar/eco-dqn}.

Graphs were generated with the NetworkX Python package~\cite{hagberg08}.  For ER graphs, a connection probability of 0.15 is used.  The BA graphs have an average degree of 4.  To produce our target graphs we then randomly set all non-zero edges to ${\pm}1$.

\subsubsection{MPNN architecture.}

The initial embedding for each vertex, $v$, is given by
\begin{equation}
	\mu_{v}^{0} = \relu(\theta_1 x_{v})
\end{equation}
where $x_{v} {\in} \R^{m}$ is the input vector of observations and $\theta_1 {\in} \R^{m \times n}$.  We also learn embeddings describing the connections to each vertex $v$,
\begin{equation}
	\xi_{v} = \relu \Big(\theta_3 \big[ \tfrac{1}{\abs{N(v)}} \sum_{u{\in}N(v)}  \relu \big( \theta_2 [w_{uv}, x_{u}] \big), \abs{N(v)} \big]  \Big),
\end{equation}
where $\theta_2 {\in} \R^{m{+}1 \times n{-}1}$, $\theta_3 {\in} \R^{n \times n}$ and square bracket denote concatenation.  The embeddings at each vertex are then updated according to
\begin{align}
	m_{v}^{k+1} &= \relu\Big( \theta_{4,k} \big[ \tfrac{1}{\abs{N(v)}} \sum_{u{\in}N(v)} w_{uv} \mu_{u}^{k}, \xi_{v} \big] \Big), \\
	\mu_{v}^{k+1} &= \relu\Big( \theta_{5,k} \big[ \mu_{v}^{k},  m_{v}^{k+1} \big] \Big),
\end{align}
where $\{ \theta_{4,k}, \theta_{5,k} \} {\in} \R^{2n \times n}$.  After $K$ rounds of message passing, the Q-value for a vertex is read out using the final embeddings across the entire graph,
\begin{equation}
	Q_{v} = \theta_{7} \Big[ \relu\big(  \theta_{6} \tfrac{1}{\abs{V}} \sum_{u{\in}V} \mu_{u}^{K} \big), \mu_{v}^{K} \Big],
\end{equation}
with $\theta_6 {\in} \R^{n \times n}$ and $\theta_7 {\in} \R^{2n}$.

In this work we use $n{=}64$ dimensional embedding vectors, and have $K{=}3$ rounds of message passing.  However, many different MPNN implementations can be used with good success. In general, what is important is that the network be capable of capturing relevant information about the local neighbourhood of a vertex.

\subsubsection{Training details.}

The Q-learning algorithm for ECO-DQN is shown in algorithm 1.  All agents are trained with a minibatch sizes of 64 and $k{=}32$ actions per step of gradient descent.  The learning rate is $10^{-4}$ and the exploration rate is linearly decreased from $\varepsilon{=}1$ to $\varepsilon{=}0.05$ over the first ${\sim}\SI{10}{\percent}$ of training.  We use the same MPNN for both S2V-DQN and ECO-DQN.  We verify that this network properly represents S2V-DQN by reproducing its performance on the `Physics' dataset at the level reported  in the original work by Khalil~\etal{}~\shortcite{khalil17}. During the training of an irreversible agent's Q-network, the predictions of the Q-values produced by the target network are clipped to be strictly non-negative.  This clipping is also used by Khalil~\etal{} and is empirically observed to improve and stabilise training.

\begin{algorithm}
	\DontPrintSemicolon
	Initialize experience replay memory $\mathcal{M}$.
	
	\For{\textup{each episode}}{
	
	Sample a graph $G(V,W)$ from distribution $\mathbb{D}$\;
	Initialise a random solution set, $S_0 \subset V$\;
	
	\For{\textup{each step }t\textup{ in the episode}}{
	$v_t = 
	\begin{cases}
   		\text{choose random }v^{\prime} \in {V},& \text{with prob.\ }\varepsilon \\
    	\mathrm{argmax}_{v^{\prime} \in V} Q(\State_t,v^{\prime};\theta),              & \text{otherwise}
	\end{cases}$\;
	$S_{t+1} :=
	\begin{cases}
   		S_t \cup \{v_t\},& \text{if }v_t \notin S_t \\
    	S_t \setminus \{v_t\},& \text{if }v_t \in S_t \\
	\end{cases}$\;
	Add tuple $(\State_t,v_t,\Reward_t,\State_{t+1})$ to $\mathcal{M}$\;
	
	\If{$t\bmod k=0$}{
	\textup{Sample minibatch} $B \sim \mathcal{M}$\;
	\textup{Update }$\theta$\textup{ by SGD for }$B$\;
	
	} 
	} 
	} 
	
	\Return $\theta$\;
    \caption{Q-learning for ECO-DQN}   
\end{algorithm}

\subsubsection{Extended data.}

ECO-DQN's generalisation performance on ER and BA graphs is shown in table \ref{tab:performance_comps_app}. ECO-DQN and S2V-DQN have a similar computational cost per time-step, which is largely determined by computing the MPNN-predicted Q-values, with the overall time to solution depending on the graph size and episode length.  This performance is compared to the speed of simple MCA agents in table~\ref{tab:performance_comps_time}.

\begin{table}[!ht]
\centering
\begin{tabular}{|c||c|c|}
\hline
Graph size & ECO-DQN & MCA-(rev,\,irrev) \\
\hline
$\abs{V}{=}20$ & \SI{0.29\pm0.06}{\ms} & \SI{0.10\pm0.02}{\ms} \\
$\abs{V}{=}500$ & \SI{6.64\pm0.09}{\ms} & \SI{1.38\pm0.06}{\ms} \\
\hline
\end{tabular}
\caption{Time per action for ECO-DQN and the greedy MCA algorithms across different graph sizes. Experiments were performed on NVIDIA Tesla M60 GPUs.}
\label{tab:performance_comps_time}
\end{table}

\FloatBarrier

\fontsize{8.5pt}{10.0pt} \selectfont

\bibliographystyle{aaai}
\bibliography{refs}

\clearpage

\section*{Supplemental Material}

\def\arraystretch{1}
\setlength\tabcolsep{3pt}

\begin{table}[!b]
\centering
\begin{tabular}{|l|c|c|c|c|c|c|}
\hline
$\indices{\mathrm{Approach}}{\abs{V}}$ & 20 & 40 & 60 & 100 & 200 & 500 \\
\hline
\rowcolor{Gray} \multicolumn{7}{|c|}{ER graphs} \\
\hline
ECO-DQN & 0.99 & 1.00 & 1.00 & 1.00 & 1.00 & 1.00 \\
S2V-DQN & 0.97 & 0.99 & 0.99 & 0.98 & 0.96 & 0.95 \\
MCA & 1.00 & 1.00 & 1.00 & 0.99 & 0.98 & 0.96 \\
CPLEX & 1.00 & 1.00 & 1.00 & 0.87 & 0.46 & 0.16 \\
SimCIM & 1.00 & 1.00 & 1.00 & 1.00 & 0.99 & 0.99 \\
Leleu~\etal{} & 1.00 & 1.00 & 1.00 & 1.00 & 1.00 & 1.00 \\
\hline
\rowcolor{Gray} \multicolumn{7}{|c|}{BA graphs} \\
\hline
ECO-DQN & 1.00 & 1.00 & 1.00 & 1.00 & 1.00 & 0.99 \\
S2V-DQN & 0.97 & 0.98 & 0.98 & 0.97 & 0.96 & 0.92 \\
MCA & 1.00 & 1.00 & 1.00 & 0.99 & 0.95 & 0.90 \\
CPLEX & 1.00 & 1.00 & 1.00 & 1.00 & 0.83 & 0.17 \\
SimCIM & 1.00 & 1.00 & 1.00 & 0.99 & 0.99 & 0.97 \\
Leleu~\etal{} & 1.00 & 1.00 & 1.00 & 1.00 & 0.94 & 1.00 \\
\hline
\end{tabular}
\label{tab:optimization_methods_performance_BA}
\caption{The approximation ratios, averaged across 100 graphs for each graph structure and size, of the different optimization methods.}
\label{tab:optimization_methods_performance}
\end{table}

We apply six different optimization methods to the 100 validation graphs of each structure (ER or BA graphs) and size ($\abs{V}{\in}\{20,40,60,100,200,500\}$).  The highest cut value across the board is then chosen as the reference point that we refer to as the ``optimum value". 
Table~\ref{tab:optimization_methods_performance} compares the performance of these methods on our validation sets.  
The number of times each method reaches these ``optimum'' solutions is then shown in table~\ref{tab:optimization_methods_opts}.
Below we give an overview of each method and summarise their efficacy.

\subsubsection{ECO-DQN.}

The framework introduced and discussed in detail in the main text.  We train distinct agents on every graph structure and size, up to $\abs{V}=200$, and then test them on graphs of the same or larger sizes. For each individual agent-graph pair, we run 50 randomly initialised optimization episodes. Therefore, graphs with $\abs{V}=20$ are subject to only 50 optimization attempts, whereas graphs with $\abs{V}=500$ are optimised with 300 episodes using 5 distinct agents.  
 The performance of each agent is summarised in table 2 of the main text.
\subsubsection{S2V-DQN.}

An RL framework for graph-based combinatorial problems introduced by Khalil~\etal{}~\shortcite{khalil17}.  Details of our implementation can, again, be found in the main text.  S2V-DQN agents are trained and tested equivalently to the ECO-DQN agents.  However, as S2V-DQN is deterministic at test time, only a single optimization episode is used for every agent-graph pair.

\subsubsection{MCA.}

The final optimization method introduced in the main text is MaxCutApprox (MCA).  This is a simple greedy algorithm that can be appliedeither in the reversible or irreversible setting.  We refer to these as MCA-rev and MCA-irrev, respectively.  For every optimization episode of ECO-DQN or S2V-DQN, a corresponding MCA-rev or MCA-irrev episode is also undertaken.  We take the best solution found in any episode by either of these greedy algorithms as the MCA solution.

We see from tables~\ref{tab:optimization_methods_performance} and~\ref{tab:optimization_methods_opts} that the greedy MCA algorithms find optimal solutions on nearly all graphs of size up to $\abs{V}=60$, but performance rapidly deteriorates thereafter.  As the number of possible solution configurations (states) grows exponentially with the number of vertices, this simply reflects how it quickly becomes intractable to sufficiently cover the state-space in our finite number of episodes.

\subsubsection{CPLEX.}

A commercial optimizer.  We first transform the Max-Cut problem into a QUBO (Quadratic Unconstrained Binary Optimization) task~\cite{kochenberger06}.  Strictly, for a graph, $G(V,W)$, with vertices $V$ connected by edges $W$, this task is to minimize the Hamiltonian
\begin{equation}
	H = - \sum_{i,j} w_{ij}(x_i - x_j)^{2},
\end{equation}
where $x_k \in \{\pm1\}$ labels whether vertex $k \in V$ is in the solution subset, $S \subset V$.  This Hamiltonian is then solved using mixed integer programming (MIP) by the CPLEX branch-and-bound routine.

For each graph, we take the best solution found within \SI{10}{\minute} as the final answer.  Within this time budget we find the exact solution on all graphs with up to 60 vertices.  Only some of the 100-vertex graphs are optimally solved, with performance significantly dropping for the 200 and 500 vertex graphs due to the unfeasibly large solution space.

\subsubsection{SimCIM.}

A simulated annealing heuristic proposed by Tiunov~\etal{}~\shortcite{tiunov19} that models the classical dynamics within a coherent Ising machine (CIM)~\cite{yamamoto17}.  This approach relaxes the binary vertex values (${x_k \in \{\pm1\}}$),  associated with labelling a vertex as either in or out of the solution subset, to analog values ($-1 \leq x_k \leq 1$).  Each vertex is initialised to 0, and then subjected to evolution according to a set of stochastic differential equations that describe the operation of the CIM. In the process of the evolution, the system eventually settles with all vertices in near-binary states. Details of both CIM and SimCIM beyond the high-level description given here can be found in the referenced works.  The hyperparameters of SimCIM were optimised using a differential evolution approach by M-LOOP~\cite{wigley16} over 50 runs.

\subsubsection{Leleu~\etal{}.}

Another recently developed simulated annealing heuristic that relaxes the binary vertex labels to analog values.  A key feature of this approach is the modification of the time-dependent interaction strengths in such a way as to destabilise locally optimal solutions.  Details can be found in the work of Leleu~\etal{}~\shortcite{leleu19}.  As with SimCIM, the hyperparameters are adjusted by M-LOOP~\cite{wigley16} over 50 runs.

\def\arraystretch{1}
\setlength\tabcolsep{3pt}

\begin{table}[!b]
\centering
\begin{tabular}{|l|c|c|c|c|c|c|}
\hline
$\indices{\mathrm{Approach}}{\abs{V}}$ & 20 & 40 & 60 & 100 & 200 & 500 \\
\hline
\rowcolor{Gray} \multicolumn{7}{|c|}{ER graphs} \\
\hline
ECO-DQN & 99 & 100 & 100 & 100 & 100 & 50 \\
S2V-DQN & 76 & 67 & 55 & 18 & 0 & 0 \\
MCA & 100 & 100 & 93 & 49 & 3 & 0 \\
CPLEX & 100 & 100 & 100 & 0 & 0 & 0 \\
SimCIM & 100 & 92 & 92 & 71 & 13 & 0 \\
Leleu~\etal{} & 100 & 100 & 100 & 98 & 100 & 92 \\
\hline
\rowcolor{Gray} \multicolumn{7}{|c|}{BA graphs} \\
\hline
ECO-DQN & 100 & 100 & 100 & 100 & 95 & 12 \\
S2V-DQN & 74 & 60 & 39 & 18 & 1 & 0 \\
MCA & 100 & 99 & 85 & 24 & 0 & 0 \\
CPLEX & 100 & 100 & 100 & 97 & 0 & 0 \\
SimCIM & 100 & 87 & 94 & 60 & 12 & 0 \\
Leleu~\etal{} & 100 & 97 & 99 & 100 & 96 & 97 \\
\hline
\end{tabular}
\label{tab:optimization_methods_opts_BA}
\caption{The relative contributions of the different optimization methods to the ``optimum'' solutions.  Shown is the number of graphs (out of 100) for which each approach finds the best, or equal best, solution.}
\label{tab:optimization_methods_opts}
\end{table}

\end{document}